\documentclass{article}
\usepackage{ijcai26}

\usepackage{times}
\usepackage{soul}
\usepackage{url}
\usepackage[hidelinks]{hyperref}
\usepackage[utf8]{inputenc}
\usepackage[small]{caption}
\usepackage{graphicx}
\usepackage{amsmath}
\usepackage{amsthm}
\usepackage{booktabs}
\usepackage{algorithm}
\usepackage{algorithmic}
\usepackage[switch]{lineno}
\usepackage{algorithm} 
\usepackage{algorithmic}
\usepackage{tcolorbox}
\usepackage{amssymb}
\usepackage{dsfont} 
\usepackage{multirow}

\usepackage{booktabs}
\usepackage{multirow}
\usepackage{colortbl}  
\usepackage{xcolor}  

\title{Online Self-Calibration Against Hallucination in Vision-Language Models}

\author{
    Minghui Chen$^{1,2}$\footnotemark[1]~\and
    Chenxu Yang$^{1,2}$\thanks{$\quad$ Equal Contribution}\and
    Hengjie Zhu$^{1,2}$\and 
    Dayan Wu$^{1,2}$\thanks{$\quad$ Dayan Wu is the corresponding author.}
    \and
    Zheng Lin$^{1,2}$\and
    Qingyi Si$^3$
\affiliations
   \textsuperscript{\rm 1}Institute of Information Engineering, Chinese Academy of Sciences, Beijing, China \\
  \textsuperscript{\rm 2}School of Cyber Security, University of Chinese Academy of Sciences, Beijing, China \\
  \textsuperscript{\rm 3}JD.COM  
  \\
}

\begin{document}

\maketitle

\begin{abstract}
    
    Large Vision-Language Models (LVLMs) often suffer from hallucinations, generating descriptions that include visual details absent from the input image. Recent preference alignment methods typically rely on supervision distilled from stronger models such as GPT. However, this offline paradigm introduces a Supervision-Perception Mismatch: the student model is forced to align with fine-grained details beyond its perceptual capacity, learning to guess rather than to see. To obtain reliable self-supervision for online learning, we identify a Generative-Discriminative Gap within LVLMs, where models exhibit higher accuracy on discriminative verification than open-ended generation. Leveraging this capability, we propose \textbf{O}nline \textbf{S}elf-\textbf{CA}lib\textbf{R}ation (OSCAR), a framework that integrates Monte Carlo Tree Search with a Dual-Granularity Reward Mechanism to construct preference data and iteratively refines the model via Direct Preference Optimization. Extensive experiments demonstrate that OSCAR achieves state-of-the-art performance on hallucination benchmarks while improving general multimodal capabilities.
\end{abstract}

\section{Introduction}

\begin{figure}[t]
    \centering
    \includegraphics[width=1.0\linewidth]{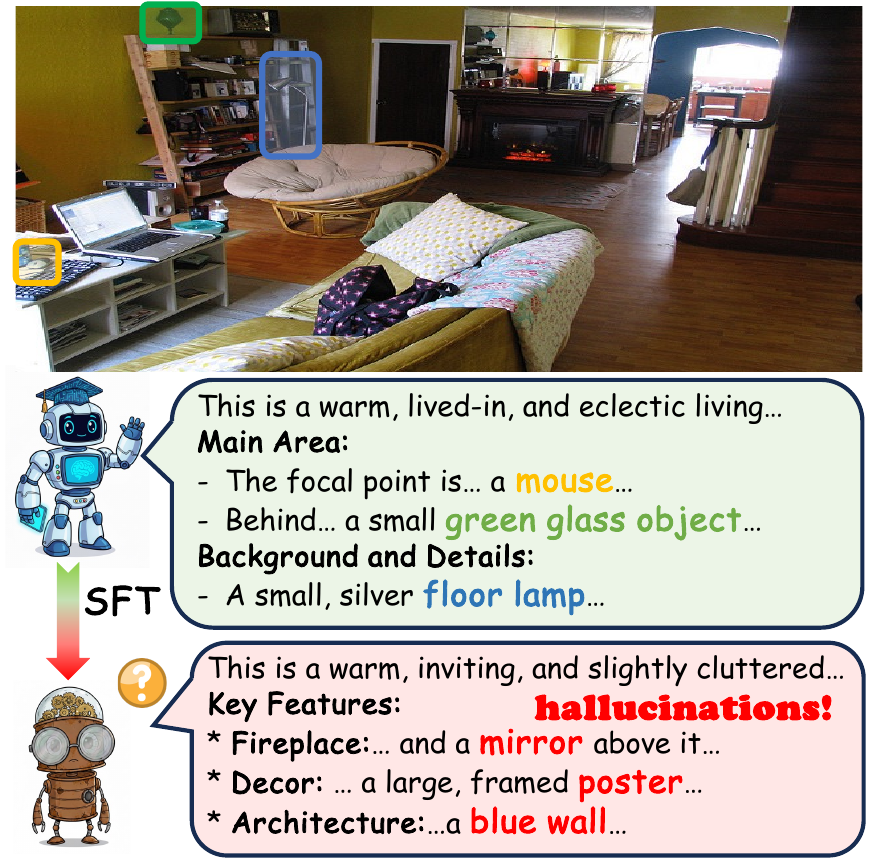}
    \caption{Supervision-Perception Mismatch. Offline supervision from a stronger teacher model forces the student to describe visual details it cannot reliably perceive, resulting in hallucinations.}
    \label{fig:perception_mismatch}
\end{figure}

Large Vision-Language Models (LVLMs)~\cite{llava-liu2023visual,minigpt-zhu2023minigpt,qwen-vl-bai2023qwen,instruct-blip-dai2023instructblip,deepseek-lu2024deepseek} integrate visual encoders with pre-trained Large Language Models (LLMs), achieving strong performance across a wide range of multimodal tasks, from image captioning to visual reasoning. However, these models frequently suffer from hallucinations~\cite{survey1-rawte2023survey,survey2-bai2024hallucination,survey3-liu2024survey}, generating content that is inconsistent with or absent from the visual input, such as fabricating non-existent objects, misinterpreting spatial relationships, or incorrectly describing object attributes. This limitation poses a significant barrier to deploying LVLMs in safety-critical domains like autonomous driving, medical imaging, and robotics, where factual grounding is essential.

Recent efforts to mitigate hallucinations have largely relied on preference alignment techniques, including Reinforcement Learning from Human Feedback (RLHF)~\cite{rlhf-sun2024aligning} and DPO~\cite{dpo-rafailov2023direct}. These methods typically construct preference datasets using human annotations~\cite{rlhf-sun2024aligning,human-anno-gunjal2024detecting} or responses distilled from stronger, proprietary models such as GPT~\cite{gpt4-achiam2023gpt,distill-liu2023aligning,silkie-li2023silkie,POVID-zhou2024aligning}.
While effective to some extent, we argue that this reliance on offline supervision introduces a fundamental Supervision-Perception Mismatch. As illustrated in Fig.~\ref{fig:perception_mismatch}, teacher models with superior visual capabilities tend to produce highly detailed descriptions that capture subtle visual elements, such as small objects or fine-grained attributes that are difficult to discern. 
When a weaker student model is supervised by such data, it is compelled to reproduce these fine-grained details that lie beyond its perceptual capacity. Unable to ground these descriptions in actual visual features, the student instead resorts to exploiting language priors and statistical shortcuts, generating different but equally ungrounded content. In essence, the model learns to guess rather than to see.
This observation is further validated by our pilot experiments in Section~\ref{3.2}. We find that fine-tuning LLaVA-1.5-7B on teacher-distilled data paradoxically increases hallucination rates, with performance degrading further as more training data is added. These findings motivate a shift toward online learning, where training data respects the model's intrinsic perceptual boundaries.
A natural question arises: can we obtain high-quality, truthful supervision from a model that is itself prone to hallucination? To address this, we identify a notable Generative-Discriminative Gap within LVLMs. Our analysis shows that while models often yield to language inertia during autoregressive generation, they exhibit considerably higher accuracy on discriminative tasks, such as verifying whether a specific object exists in an image. This gap arises because discriminative verification, by explicitly conditioning on a specific query, reduces the influence of unconstrained language priors that dominate open-ended generation. This finding suggests that LVLMs possess a latent capacity for self-verification that remains underutilized during generation.

While the Generative-Discriminative Gap addresses the question of where to obtain reliable supervision, another key challenge lies in how to construct high-quality training data. Standard decoding strategies like greedy or beam search optimize locally, presenting two limitations. First, some tokens that appear safe at the current step may carry high risk of inducing hallucinations in subsequent generation, a cascading effect that local optimization cannot foresee. Second, greedily selecting branches with the lowest immediate hallucination rate at each step may compromise overall response quality in terms of logical consistency and fluency. To overcome these limitations, we integrate Monte Carlo Tree Search (MCTS) to explore the generation space more strategically~\cite{mcts1-xie2024monte,mcts2-tian2024toward,mcts3-zhang2024accessing}. We design a Dual-Granularity Reward Mechanism to guide the search. At the node level, we leverage the model's discriminative capability by prompting it to verify whether each generated sentence mentions objects absent from the image, using the probability of a negative response as the process reward. At the trajectory level, we employ a Gated Outcome Reward that evaluates response quality only if the complete response passes a faithfulness check, and returns zero otherwise.

Through MCTS backpropagation, terminal rewards propagate from leaf nodes to the root, enabling the model to identify generation trajectories that balance visual faithfulness with descriptive richness. Building on these insights, we propose \textbf{O}nline \textbf{S}elf-\textbf{CA}lib\textbf{R}ation (\textbf{OSCAR}), a unified framework for online preference learning. Specifically, OSCAR extracts preference pairs from the MCTS tree at two granularities: global path comparison selects complete trajectories with the highest and lowest cumulative values, while sibling comparison pairs nodes along the optimal path with their worst-performing siblings. These preference pairs are then used to update the model via DPO~\cite{dpo-rafailov2023direct}. At each iteration, the updated model generates new preference data through MCTS, ensuring the training distribution evolves alongside the model's capabilities and enabling continuous self-improvement.
Our contributions are as follows:
\begin{itemize}
\item 
We empirically demonstrate the necessity of online preference learning that respects the model's intrinsic perceptual boundaries, showing that offline distillation from stronger teachers can unexpectedly exacerbate hallucinations.

\item 
We propose OSCAR, a novel training paradigm that exploits the Generative-Discriminative Gap. By integrating MCTS with a Dual-Granularity Reward Mechanism, OSCAR enables lookahead to suppress early tokens that risk inducing downstream hallucinations.

\item 
Extensive experiments show that OSCAR achieves state-of-the-art performance on hallucination benchmarks while simultaneously improving general multimodal capabilities.

\end{itemize}

\section{Related Work}

\subsection{Hallucination in LVLMs}
Large Vision-Language Models (LVLMs) frequently suffer from hallucinations, generating content that is inconsistent with the visual input~\cite{survey1-rawte2023survey,survey2-bai2024hallucination}. Various approaches have been proposed to mitigate this issue, including enhancing dataset quality~\cite{related11-liu2023mitigating,human-anno-gunjal2024detecting,si-etal-2023-combo,silkie-li2023silkie}, manipulating the decoding process~\cite{related12-leng2024mitigating,related13-huang2024opera,yang2025breakingtradeofffaithfulnessexpressiveness,octopus-suo2025octopus,si-etal-2021-check}, leveraging external models for post-hoc correction~\cite{related14-yin2024woodpecker,POVID-zhou2024aligning,si-etal-2022-language}, and preference optimization~\cite{related15-xie2024v,SIMA-wang2025enhancing,yang-etal-2025-weights,si-etal-2022-towards}. Though effective to some extent, these methods predominantly rely on offline supervision, which may introduce a Supervision-Perception Mismatch when the target model lacks the perceptual capabilities to ground such fine-grained details.

\subsection{Self-Improvement for LLMs}


Self-improvement methods enable models to enhance their 
capabilities using self-generated feedback, reducing reliance 
on external annotations~\cite{related21-huang2023large,related22-yuan2024self,feedback-hu2024teaching,yang2026selfdistilledrlvr,paradox-sun2025self,dai2025sgrpoearlyexitreinforcement}. In the vision-language domain, recent works such as STIC~\cite{STIC-deng2024enhancing} and SIMA~\cite{SIMA-wang2025enhancing} have explored 
self-improvement for hallucination mitigation. However, these 
methods typically construct preference data via simple sampling 
or beam search, which fails to account for the cascading nature 
of hallucinations. We instead integrate MCTS with a Dual-Granularity Reward Mechanism, enabling lookahead to suppress tokens that risk inducing downstream hallucinations.

\subsection{Monte Carlo Tree Search}
Monte Carlo Tree Search (MCTS) has emerged as a powerful technique for enhancing reasoning in LLMs, inspired by its success in game-playing agents~\cite{related31-silver2016mastering}. Recent works have adapted MCTS to guide text generation by simulating future trajectories and backpropagating rewards, achieving improvements in mathematical reasoning~\cite{mcts1-xie2024monte,mcts2-tian2024toward,mcts3-zhang2024accessing,yang2025testtimepromptintervention,yang2025dynamicearlyexitreasoning} and task planning~\cite{related33-hao2023reasoning,related32-li2025think,yang2026system,planning-li2025chatsop}. In the context of hallucination mitigation, we are the first to leverage MCTS for preference data construction, enabling lookahead to suppress locally plausible tokens that risk inducing downstream hallucinations.


\section{Observations and Motivations}
\label{sec:preliminary}

\begin{figure}[t]
    \centering
    \includegraphics[width=1.0\linewidth]{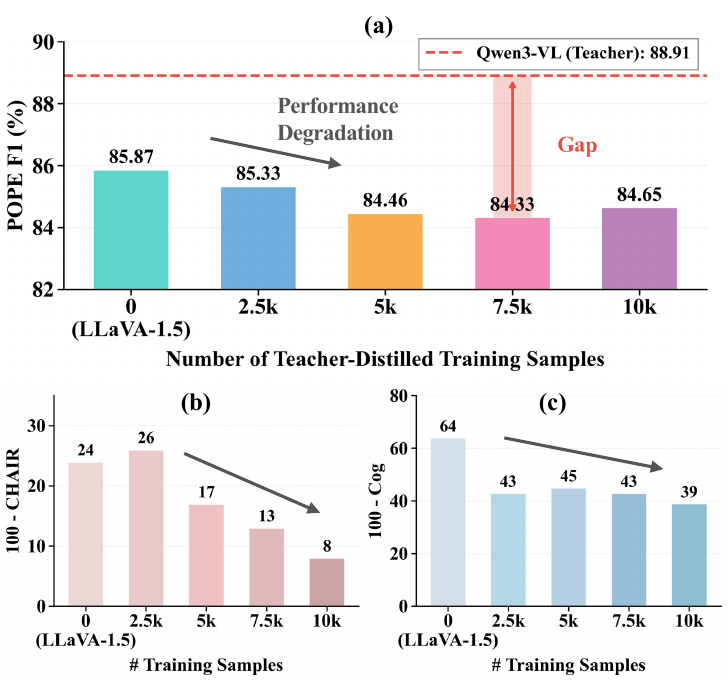}
    \caption{Performance comparison before and after supervised fine-tuning with stronger teacher-distilled data.}
    \label{fig:pre_3_2}
\end{figure}

\subsection{Supervision-Perception Mismatch} \label{3.2}

Recent preference alignment methods predominantly rely on supervision signals distilled from stronger, proprietary models such as GPT~\cite{gpt4-achiam2023gpt}. We hypothesize that this reliance on offline supervision introduces a fundamental \textit{Supervision-Perception Mismatch}: when a target model with limited visual perception capacity is trained on data generated by a teacher model with superior perceptual abilities, the student is compelled to align with fine-grained visual details that exceed its intrinsic perceptual capabilities. Consequently, the model may learn to minimize training loss by exploiting language priors and statistical shortcuts rather than grounding its outputs in visual features. As illustrated in Fig.~\ref{fig:perception_mismatch}, teacher models with superior visual capabilities tend to capture subtle visual elements that weaker models cannot reliably perceive. For instance, the teacher model identifies a "mouse" near the laptop, a "green glass object" in the corner, and a "floor lamp" in the background. When supervised by such data, the student model fails to ground these details in actual visual features, instead generating different but equally ungrounded content, such as hallucinating a "mirror" above the fireplace, a "poster", and a "blue wall" that does not exist.

To empirically validate this hypothesis, we employ Qwen3-VL-8B-Instruct~\cite{qwen-yang2025qwen3} to generate detailed image descriptions on randomly sampled images from the LLaVA-150k dataset, and use these teacher-generated descriptions to fine-tune LLaVA-1.5-7B with varying data scales (2.5k, 5k, 7.5k, and 10k samples). As shown in Fig.~\ref{fig:pre_3_2} (a), despite Qwen3-VL achieving 88.91\% on POPE F1, the fine-tuned LLaVA models consistently underperform the original baseline (85.87\%), with performance degrading further as more training data is added. Similar trends are observed on the AMBER benchmark, where both CHAIR and Cog scores deteriorate after fine-tuning (Fig.~\ref{fig:pre_3_2}b-c). This counterintuitive finding confirms that \textit{offline} supervision can paradoxically exacerbate hallucinations, motivating our shift toward \textit{online} learning.

\begin{figure}[t]
    \centering
    \includegraphics[width=1\linewidth]{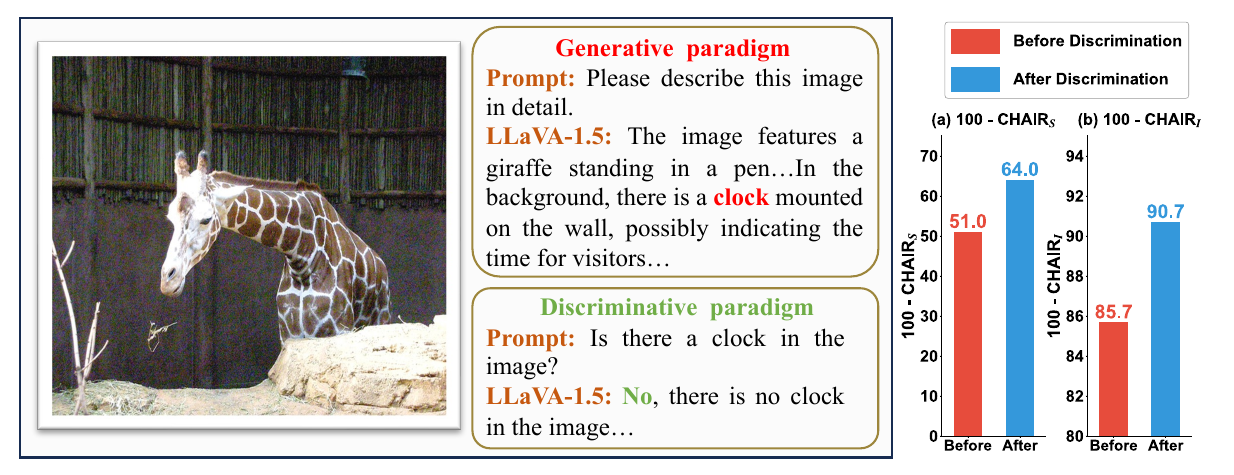}
    \caption{Illustration of the Generative-Discriminative Gap.}
    \label{fig:case-pre}
\end{figure}

\subsection{Generative-Discriminative Gap} \label{3.3}

To explore whether a hallucination-prone model can still provide useful self-supervision, we investigate its behavior across different inference paradigms. Our analysis reveals a notable \textit{Generative-Discriminative Gap}: while models frequently yield to language inertia during autoregressive generation, where plausible linguistic patterns overshadow visual grounding, they exhibit improved accuracy when tasked with discriminative verification against visual evidence. Fig.~\ref{fig:case-pre} illustrates this gap with a concrete example. When prompted to describe an image in detail, LLaVA-1.5 hallucinates a "clock mounted on the wall" that does not exist. However, when the same model is explicitly asked "Is there a clock in the image?", it correctly answers ``No''. This discrepancy suggests that discriminative verification, by conditioning on a specific query, reduces the influence of unconstrained language priors that dominate open-ended generation.

To quantify this gap, we conduct the following analysis. We first generate image captions using LLaVA-1.5-7B on 500 randomly sampled images from the COCO dataset and compute the CHAIR metrics. For each hallucinated object $x$ detected in the generated captions, we construct a discriminative query: ``Is there a/an $x$ in the image?'' and prompt the same model to respond with ``Yes'' or ``No''. If the model correctly answers ``No'', we remove this hallucinated object from the caption and recompute the CHAIR metrics. As shown in Fig.~\ref{fig:case-pre}, this simple self-verification procedure reduces CHAIR$_S$ from 49.0\% to 36.0\% and CHAIR$_I$ from 14.3\% to 9.3\%, confirming that LVLMs possess a latent capacity for self-verification that remains underutilized during standard generation. This finding motivates our approach: by leveraging the model's discriminative ability to curate training data, we can align the granularity of generated descriptions with the model's intrinsic perceptual capabilities while improving factual accuracy.

\begin{figure*}[htbp]
  \centerline{\includegraphics[scale=0.47]{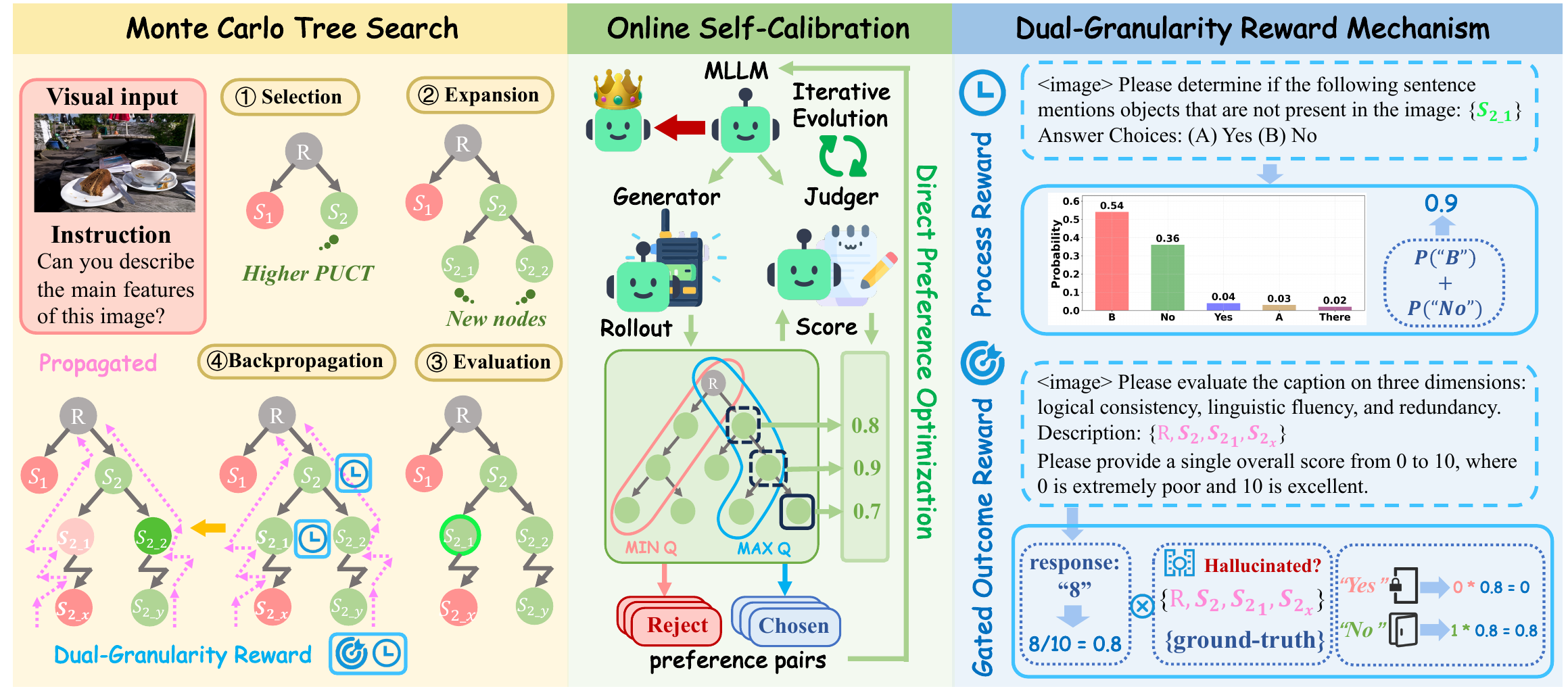}}
  \caption{Overview of Online Self-CAlibRation (OSCAR). Left: Monte Carlo Tree Search explores the generation space through selection, expansion, evaluation, and backpropagation. Middle: Preference pairs are extracted from the MCTS tree to refine the model via Direct Preference Optimization. Right: The Dual-Granularity Reward Mechanism combines a process reward that verifies each sentence, and a gated outcome reward that evaluates response quality only when the trajectory is hallucination-free.}
  \label{fighuman}
\end{figure*}

\section{Methodology}
Building upon the insights from Section~\ref{sec:preliminary}, we present \textbf{O}nline \textbf{S}elf-\textbf{CA}lib\textbf{R}ation (OSCAR) , a framework that leverages the model's capability to construct high-quality online preference data and iteratively refines the model through DPO. In this section, we first detail our MCTS-guided generation with a dual-granularity reward mechanism (\S\ref{sec:mcts}), and then describe the preference learning procedure (\S\ref{sec:preference}).

\subsection{LVLM Inference} 
A Large Vision-Language Model (LVLM) $\mathcal{M}_\theta$ with parameters $\theta$ takes as input a visual image $\mathbf{v}$ and a textual instruction $\mathbf{q}$, and generates a textual response $\mathbf{y} = (y_1, y_2, \ldots, y_T)$ in an autoregressive manner. Specifically, at each generation step $t$, the model computes the probability distribution over the vocabulary conditioned on the visual input, the textual instruction, and the previously generated tokens:
\begin{equation}
    p(y_t | \mathbf{v}, \mathbf{q}, y_{<t}; \theta) = \text{Softmax}(\ell_t),
\end{equation}
where $\ell_t$ denotes the logit vector for the next token $y_t$, and $y_{<t} = (y_1, \ldots, y_{t-1})$ represents the sequence of tokens generated before step $t$. The complete response is generated by sequentially sampling tokens from this distribution until the end-of-sequence token is produced.

\subsection{MCTS-Guided Generation}
\label{sec:mcts}

A key challenge in constructing faithful training data is that standard decoding strategies such as greedy or beam search optimize locally, which suffers from two limitations. First, tokens that seem acceptable at the current step may induce hallucinations in later generation, a long-term risk that local optimization cannot anticipate. Second, prioritizing branches with minimal immediate hallucination at each step may degrade overall response quality, sacrificing logical consistency and fluency. To overcome these limitations, we integrate Monte Carlo Tree Search (MCTS) into the generation process. By simulating future trajectories and backpropagating terminal rewards, MCTS enables the model to evaluate the long-term value of each token, identifying generation paths that balance visual faithfulness with descriptive richness.

\paragraph{MCTS Procedure.}
We decompose the generation process into sentence-level steps, where each node in the search tree represents a partial response. Formally, let $s_t = (\mathbf{v}, \mathbf{q}, a_1, a_2, \ldots, a_{t-1})$ denote the state at step $t$, where $a_i$ represents the $i$-th generated sentence. An action $a_t$ corresponds to generating a complete sentence, delimited by terminal punctuation marks. Each MCTS iteration consists of four phases:

\begin{itemize}
    \item \textbf{Selection.} Starting from the root node, we traverse the tree by selecting child nodes according to the PUCT (Predictor + Upper Confidence bounds applied to Trees) criterion~\cite{puct-rosin2011multi}:
\begin{equation}
    a^* = \arg\max_{a} \left[ Q(s, a) + c_{\text{puct}} \cdot p(a|s) \cdot \frac{\sqrt{N(s)}}{1 + N(s, a)} \right],
\end{equation}
where $Q(s, a)$ denotes the action value, $p(a|s) = \pi_\theta(a|s) / |a|^\lambda$ is the length-normalized policy probability with penalty $\lambda$, $N(s)$ and $N(s,a)$ are visit counts, and $c_{\text{puct}}$ controls the exploration-exploitation trade-off.

    \item \textbf{Expansion.} At a leaf node, we sample $K$ candidate sentences from the policy $\pi_\theta$ using temperature sampling. To ensure diversity, we filter candidates with embedding similarity exceeding a threshold $\tau_{\text{sim}}$.

    \item \textbf{Evaluation.} Each expanded node receives a value score through our \textit{Dual-Granularity Reward Mechanism}, detailed below.

    \item \textbf{Backpropagation.} After evaluation, statistics are propagated from the leaf node back to the root. The Q-value and state value are updated as:
\begin{align}
    Q(s_t, a) &= r(s_t, a) + \gamma \cdot V(s_{t+1}), \\
    V(s_t) &= \frac{\sum_{a} N(s_{t+1}) \cdot Q(s_t, a)}{\sum_{a} N(s_{t+1})},
\end{align}
where $r(s_t, a) = \text{value}(s_{t+1}) - \text{value}(s_t)$ represents the immediate reward and $\gamma$ is the discount factor.
\end{itemize}

\paragraph{Dual-Granularity Reward Mechanism.}
\label{sec:reward}

Central to our approach is a reward mechanism that combines node-level process supervision with trajectory-level outcome evaluation. This design enables the search to identify early tokens that, while locally plausible, carry high risk of inducing downstream hallucinations.

\textit{Process Reward (Node-Level).} For each generated sentence $a_t$, we employ the model's discriminative capability to assess whether the sentence mentions objects absent from the image. Specifically, we construct a verification prompt:
\begin{tcolorbox}[colback=gray!5, colframe=gray!50, boxrule=0.5pt, arc=2pt]
\small
\texttt{<image>} Please determine if the following sentence mentions objects that are not present in the image: \texttt{\{sentence\}} \\
Answer Choices: (A) Yes \quad (B) No
\end{tcolorbox}

\noindent The process reward $r_{\text{proc}}$ is computed as the probability that the model judges the sentence as hallucination-free:
\begin{equation}
    r_{\text{proc}}(a_t) = p_\theta\bigl(\text{``No''} \mid \mathbf{v}, \mathcal{P}_{\text{proc}}(a_t)\bigr),
\end{equation}
where $\mathcal{P}_{\text{proc}}(a_t)$ denotes the verification prompt instantiated with the candidate sentence $a_t$.

\textit{Gated Outcome Reward (Trajectory-Level).} To evaluate the quality of a complete trajectory, we perform a greedy rollout from the current state to a terminal state, yielding a complete response $\mathbf{y}_{\text{rollout}}$. The outcome reward incorporates a \textit{gating mechanism} that enforces strict faithfulness requirements. First, we assess whether the complete response contains any hallucinated content through a faithfulness check. Specifically, we extract all objects mentioned in the generated description and compare them against the ground-truth objects provided by the dataset. Specifically, we extract all object nouns from the generated description and map them to canonical COCO category names using a predefined synonym dictionary. If any mapped object does not appear in the ground-truth object set, the response is considered to contain hallucinations. The gating function is defined as:
\begin{equation}
    g(\mathbf{y}_{\text{rollout}}) = \mathds{1}[\mathcal{O}(\mathbf{y}_{\text{rollout}}) \subseteq \mathcal{O}_{\text{gt}}],
\end{equation}
where $\mathcal{O}(\mathbf{y}_{\text{rollout}})$ denotes the set of canonical object names extracted from the generated response, and $\mathcal{O}_{\text{gt}}$ denotes the ground-truth object set.
Second, for trajectories that pass the gate, we assess the response quality along dimensions of logical consistency, linguistic fluency, and redundancy:
\begin{tcolorbox}[colback=gray!5, colframe=gray!50, boxrule=0.5pt, arc=2pt]
\small
\texttt{<image>} Please evaluate the following caption on three dimensions: logical consistency, linguistic fluency, and redundancy. \\
Caption: \texttt{\{$\mathbf{y}_{\text{rollout}}$\}} \\
Please provide a single overall score from 0 to 10, where 0 is extremely poor and 10 is excellent.
\end{tcolorbox}
\noindent Let $\text{score}_{\text{quality}} \in [0, 10]$ denote the extracted quality score. The gated outcome reward is defined as:
\begin{equation}
    r_{\text{out}}(\mathbf{y}_{\text{rollout}}) = 
    \begin{cases}
        \text{score}_{\text{quality}} / 10 & \text{if } g(\mathbf{y}_{\text{rollout}}) = 1, \\
        0 & \text{otherwise}.
    \end{cases}
\end{equation}

\begin{table*}[t]
\centering
\small
\resizebox{0.98\textwidth}{!}{
\begin{tabular}{l cc cccc c cc c}
\toprule
\multirow{3}{*}{\textbf{Method}} &\multicolumn{7}{c}{\textbf{Generative Task}} &\multicolumn{3}{c}{\textbf{Discriminative Task}} \\ 
\cmidrule(lr){2-8} \cmidrule(lr){9-11}
&\multicolumn{2}{c}{\textbf{Object HalBench}} &\multicolumn{4}{c}{\textbf{AMBER-Gen.}} &{\textbf{MM-VET}} &\multicolumn{2}{c}{\textbf{AMBER-Dis.}} &\textbf{POPE} \\
\cmidrule(lr){2-3} \cmidrule(lr){4-7} \cmidrule(lr){8-8}\cmidrule(lr){9-10} \cmidrule(lr){11-11}
&CHAIR$_{s}\downarrow$ &CHAIR$_{i}\downarrow$ &CHAIR$\downarrow$ &Cover$\uparrow$ &Hal$\downarrow$ &Cog$\downarrow$ &Overall$\uparrow$&Acc$\uparrow$ &F1$\uparrow$ &F1$\uparrow$ \\
\midrule
InstructBLIP\cite{instruct-blip-dai2023instructblip}& 42.3 &\textbf{7.9} & 9.4 & 51.8 & 40.5 & 4.8 & 25.7 & 74.3 & \underline{79.9} & 78.56 \\
MiniGPT-4\cite{minigpt-zhu2023minigpt} & 31.4 & 11.1 & 13.6 & \textbf{63.0} & 65.3 & 11.3 & 22.1 & 63.6 & 64.7 &  61.50\\
mPLUG-Owl2\cite{mplug-ye2024mplug} & 54.4 & 12.0 & 10.6 & \underline{52.2} & 39.9 & 4.5 & \textbf{37.6} & \underline{75.6} & 78.5 & 86.20 \\
\midrule
LLaVA-1.5-7B~\cite{llava-liu2023visual}
& 49.0 & 14.3 & 7.6 & 49.6 & 31.2 & 3.6 & 32.5 & 72.2 & 75.5 & 85.87 \\
\quad +STIC~\cite{STIC-deng2024enhancing}
& - & - & 7.6 & 52.1 & 35.8 & 4.4 & 31.8 & 71.6 & 74.2 & - \\
\quad +POVID~\cite{POVID-zhou2024aligning}
& 33.6 & 9.0 & \underline{5.0} &  50.1 & 28.6 & 3.0 & 31.7 & 71.9 & 74.7 & \textbf{86.90} \\
\quad +SIMA~\cite{SIMA-wang2025enhancing}
& 40.9 & 10.4 & 6.4 & 47.4 & 26.1 & 3.2 & 31.6 & 73.4 & 76.4 & - \\
\quad +Self-Rewarding
& 38.4 & 11.2 & 6.8 & 48.2 & 27.5 & 3.0 & 32.8 & 73.1 & 76.8 & 85.93 \\
\quad +\textbf{OSCAR (Iter1)}
& 32.0 & 9.7 & 5.6 & 46.0 & 22.1 & 2.1 & 32.9 & 74.4 & 78.3 & 86.04 \\
\quad +\textbf{OSCAR (Iter2)}
& \underline{28.6} & 9.0 & 5.1 & 45.4 & \underline{19.4} & \underline{1.9} & 33.8 & 75.3 & 79.4 & 86.07 \\
\rowcolor{gray!10}
\quad +\textbf{OSCAR (Iter3)}
& \textbf{27.6} & \underline{8.2} & \textbf{4.5} & 45.7 & \textbf{17.2} & \textbf{1.6} & \underline{34.6} & \textbf{75.8} & \textbf{80.2} & \underline{86.22} \\
\midrule
LLaVA-1.5-13B~\cite{llava-liu2023visual}
& 44.8 & 11.8 & 6.4 & \textbf{50.9} & 30.3 & 3.1 & \textbf{37.6} & 67.9 & 69.1 & 86.67 \\
\quad +Self-Rewarding
& 35.2 & 9.6 & 5.4 & 48.3 & 24.6 & 2.4 & 36.8 & 69.5 & 71.2 & 86.78 \\
\quad +\textbf{OSCAR (Iter1)}
& 16.4 & 5.8 & 3.3 & 45.8 & 13.3 & 0.9 & \underline{37.4} & \underline{71.9} & \underline{74.1} & 86.89 \\
\quad +\textbf{OSCAR (Iter2)}
& \underline{7.8} & \underline{2.8} & \underline{2.8} & \underline{48.7} & \underline{9.9} & \underline{0.6} & 35.6 & 70.3 & 71.9 & \underline{87.20} \\
\rowcolor{gray!10}
\quad +\textbf{OSCAR (Iter3)}
& \textbf{5.4} & \textbf{2.6} & \textbf{2.6} & 47.6 & \textbf{8.0} & \textbf{0.5} & 36.5 & \textbf{72.4} & \textbf{74.6} & \textbf{87.26} \\
\bottomrule
\end{tabular}
}
\caption{Comparison with state-of-the-art methods on hallucination benchmarks. We evaluate on both generative tasks and discriminative tasks. The best results are shown in \textbf{bold} and the second best results are \underline{underlined}. $\downarrow$ indicates lower is better, $\uparrow$ indicates higher is better.}
\label{tab:main_results}
\end{table*}

\noindent The final value assigned combines both granularities:
\begin{equation}
    \text{value}(s_t, a_t) = r_{\text{proc}}(s_t, a_t) + r_{\text{out}}(\mathbf{y}_{\text{rollout}}).
\end{equation}
Through MCTS backpropagation, this trajectory-level reward signal propagates from leaf nodes to the root, elevating the estimated value of early tokens that lead to faithful and high-quality completions.

\subsection{Iterative Preference Learning}
\label{sec:preference}

\paragraph{Preference Pair Extraction.}
We extract preference pairs from the MCTS tree at two levels of granularity. For \textit{global path comparison}, we identify the complete path with the highest cumulative Q-value as the chosen response $\mathbf{y}^+$ and the one with the lowest Q-value as the rejected response $\mathbf{y}^-$:
\begin{equation}
    \mathbf{y}^+ = \arg\max_{\mathbf{y} \in \mathcal{T}} Q(\mathbf{y}), \quad \mathbf{y}^- = \arg\min_{\mathbf{y} \in \mathcal{T}} Q(\mathbf{y}),
\end{equation}
where $\mathcal{T}$ denotes the set of all complete trajectories in the tree. For \textit{sibling comparison}, we traverse each depth along the optimal path and pair the selected node with its worst-performing sibling, if their Q-value difference exceeds a threshold $\delta_Q$:
\begin{equation}
\resizebox{0.91\linewidth}{!}{$
    (\mathbf{y}^+_d, \mathbf{y}^-_d) = (s_{<d} \oplus a_d^*, \; s_{<d} \oplus a_d^{\text{worst}}), \; \text{if } Q(a_d^*) - Q(a_d^{\text{worst}}) \geq \delta_Q
    $},
\end{equation}
where $s_{<d}$ denotes the partial response up to depth $d$, and $\oplus$ denotes concatenation. This step-wise comparison enables the extraction of multiple preference pairs from a single MCTS tree, maximizing the utilization of information accumulated during the search process.

\paragraph{DPO Training.}
Given the preference dataset $\mathcal{D} = \{(\mathbf{v}_i, \mathbf{q}_i, \mathbf{y}_i^+, \mathbf{y}_i^-)\}_{i=1}^N$ constructed via MCTS, we update the model using Direct Preference Optimization (DPO)~\cite{dpo-rafailov2023direct}. The DPO objective directly optimizes the policy to prefer chosen responses over rejected ones:
\begin{equation}
    \mathcal{L}_{\text{DPO}}(\theta) = -\mathbb{E}_{(\mathbf{v}, \mathbf{q}, \mathbf{y}^+, \mathbf{y}^-) \sim \mathcal{D}} \left[ \log \sigma \left( \beta \cdot h_\theta(\mathbf{y}^+, \mathbf{y}^-) \right) \right],
\end{equation}
where $\sigma(\cdot)$ is the sigmoid function, $\beta$ is a temperature parameter, and:
\begin{equation}
    h_\theta(\mathbf{y}^+, \mathbf{y}^-) = \log \frac{\pi_\theta(\mathbf{y}^+ | \mathbf{v}, \mathbf{q})}{\pi_{\text{ref}}(\mathbf{y}^+ | \mathbf{v}, \mathbf{q})} - \log \frac{\pi_\theta(\mathbf{y}^- | \mathbf{v}, \mathbf{q})}{\pi_{\text{ref}}(\mathbf{y}^- | \mathbf{v}, \mathbf{q})}.
\end{equation}

\noindent Here, $\pi_{\text{ref}}$ denotes the reference policy, initialized as the model checkpoint before the current iteration. The training proceeds iteratively: at each iteration $m$, we use the current policy $\pi_\theta^{(m)}$ to construct new preference data via MCTS, then update the model to obtain $\pi_\theta^{(m+1)}$. This online paradigm ensures that the training distribution evolves alongside the model's improving capabilities, progressively tightening the alignment between generated content and the model's perceptual boundaries.

\section{Experiments}

\subsection{Experimental Setup}

\paragraph{Evaluation Benchmarks.} We conduct evaluations on both generative and discriminative hallucination tasks. For the generative task, we evaluate on Object-HalBench~\cite{chair-rohrbach2018object}, AMBER~\cite{amber-wang2023llm}, and MM-VET~\cite{mmvet-yu2023mm}. For the discriminative task, we report results on AMBER~\cite{amber-wang2023llm} and POPE~\cite{pope-li2023evaluating}. Detailed descriptions of evaluation metrics and benchmark designs are provided in the Appendix.

\paragraph{Baselines.} We compare OSCAR with three categories of methods: (1) other open-source LVLMs, including InstructBLIP~\cite{instruct-blip-dai2023instructblip}, MiniGPT-4~\cite{minigpt-zhu2023minigpt}, and mPLUG-Owl2~\cite{mplug-ye2024mplug}; (2) SoTA data-driven preference learning methods for hallucination mitigation, including STIC~\cite{STIC-deng2024enhancing}, POVID~\cite{POVID-zhou2024aligning}, and SIMA~\cite{SIMA-wang2025enhancing}; (3) a Self-Rewarding baseline that employs the same hallucination detection reward but constructs preference data via beam search instead of MCTS.

\paragraph{Implementation Details.} We adopt LLaVA-1.5-7B and LLaVA-1.5-13B as base models. For preference data construction, we sample images and prompts from LLaVA-150k~\cite{llava-liu2023visual}, yielding 120k preference pairs per iteration. The MCTS search is configured with $c_{\text{puct}}=1.0$, length penalty $\lambda=1.25$, and $Q$-value difference threshold $\delta_Q=0.05$. For DPO training, we employ LoRA~\cite{lora-hu2022lora} with rank 128 and $\alpha=256$, using the Adam optimizer with a learning rate of $1 \times 10^{-5}$ and temperature $\beta=0.1$. The model is iteratively trained for 3 iterations.

\subsection{Main Results}

\textbf{Overall Performance.} Tab.~\ref{tab:main_results} presents comprehensive comparisons between OSCAR and existing methods on hallucination benchmarks. Our method achieves state-of-the-art performance on both generative and discriminative tasks, demonstrating its effectiveness in mitigating hallucinations while preserving general multimodal capabilities.

\noindent\textbf{Generative Task.}
On Object-HalBench, OSCAR substantially reduces hallucination metrics for LLaVA-1.5-7B, decreasing CHAIR$_S$ from 49.0 to 27.6 and CHAIR$_I$ from 14.3 to 8.2. These improvements significantly surpass prior methods such as POVID (33.6/9.0) and SIMA (40.9/10.4). On AMBER generative metrics, OSCAR achieves the lowest Hal score (17.2) and Cog score (1.6), indicating fewer hallucinated responses and reduced reliance on human-like cognitive shortcuts. Notably, on MM-VET, which evaluates general multimodal understanding, OSCAR improves the overall score from 32.5 to 34.6, demonstrating that our hallucination mitigation does not compromise response quality or descriptive richness. For the larger LLaVA-1.5-13B model, OSCAR yields even more substantial gains, achieving CHAIR$_S$ of 5.4 and CHAIR$_I$ of 2.6, which represent reductions of 87.9\% and 78.0\% respectively compared to the baseline.

\noindent\textbf{Discriminative Task.} 
On discriminative benchmarks, OSCAR also demonstrates consistent improvements. For AMBER discrimination, OSCAR improves accuracy from 72.2\% to 75.8\% and F1 score from 75.5\% to 80.2\% on LLaVA-1.5-7B. On POPE, OSCAR achieves an F1 score of 86.22\%, comparable to POVID (86.90\%), while significantly outperforming it on generative tasks. These results confirm that OSCAR enhances the model's visual grounding capability across both generative and discriminative paradigms.

\noindent\textbf{Iterative Improvement.}
A key advantage of our approach is its ability to enable continuous self-improvement through iterative training. As shown in Tab.~\ref{tab:main_results}, performance improves progressively from Iter1 to Iter3 across all metrics. For LLaVA-1.5-7B, CHAIR$_S$ decreases from 32.0 (Iter1) to 28.6 (Iter2) and further to 27.6 (Iter3), while the AMBER Hal score drops from 22.1 to 19.4 to 17.2 over the same iterations. Similar trends are observed for LLaVA-1.5-13B, where CHAIR$_S$ decreases from 16.4 (Iter1) to 7.8 (Iter2) and finally to 5.4 (Iter3). This progressive enhancement validates our online learning paradigm: as the model improves, the quality of MCTS-generated preference data also increases, creating a virtuous cycle that enables sustained capability growth.

\subsection{Analysis}

\paragraph{Ablation Studies}
To analyze the contribution of each component in OSCAR, we conduct ablation studies on LLaVA-1.5-7B using a single iteration. Results are shown in Tab.~\ref{tab:ablation}. \textbf{Effect of Process Reward.} Comparing Index 4 and 5, adding the process reward substantially reduces CHAIR$_S$ from 44.0 to 32.0, demonstrating that node-level hallucination feedback is essential for fine-grained guidance during tree search. \textbf{Effect of Gated Outcome Reward.} Comparing Index 3 and 5, incorporating the gated outcome reward reduces CHAIR$_S$ from 45.6 to 32.0, ensuring trajectory-level faithfulness that complements the local process reward. \textbf{Effect of MCTS.} Comparing Index 2 and 5, integrating MCTS dramatically reduces CHAIR$_S$ from 46.7 to 32.0, confirming that lookahead search is crucial for identifying tokens that may induce downstream hallucinations. The full model significantly outperforms all partial configurations, indicating that the three components work synergistically.

\begin{table}
    \centering
    \renewcommand{\arraystretch}{1.2} 
    \setlength{\tabcolsep}{0pt} 
    
    \newcolumntype{I}{>{\centering\arraybackslash}p{3.2em}}
    \newcolumntype{W}{>{\centering\arraybackslash}p{3.15em}} 
    \newcolumntype{C}{>{\centering\arraybackslash}p{3.6em}} 
    
    \begin{tabular}{I|W W W|C C C}
        \hline
        Index & PR & GOR  & MCTS &  CHAIR$_{s}$&  CHAIR$_{i}$& POPE\\
        \hline
        1 & - & - & - & 49.0 & 14.3 & 85.87  \\
        2 & \checkmark & - & - & 46.7 & 13.8 & 86.01  \\
        3 & \checkmark & - & \checkmark & 45.6  & 13.5 & 86.00    \\
        4 & - & \checkmark & \checkmark & 44.0 & 12.6 & 86.03  \\
        5 & \checkmark & \checkmark  & \checkmark  & 32.0& 9.7 & 86.04\\
        \hline
    \end{tabular}
    
    \caption{Ablation study on key components of OSCAR. PR: Process Reward; GOR: Gated Outcome Reward.}
    \label{tab:ablation}
\end{table}

\begin{table}[t]
\centering
\renewcommand{\arraystretch}{1.1}
\begin{tabular}{lccc}
\toprule
\textbf{Method} & \textbf{CHAIR$\downarrow$} & \textbf{Hal$\downarrow$} & \textbf{Cog$\downarrow$} \\
\midrule
LLaVA-1.5-7B & 7.6 & 31.2 & 3.6 \\
\midrule
+SFT (Qwen3-VL) & 9.2 & 62.7 & 6.1 \\
+SFT (LLaVA) & 7.5 & 30.6 & 3.4 \\
+SFT (OSCAR) & \textbf{4.5} & \textbf{15.4} & \textbf{1.4} \\
\bottomrule
\end{tabular}
\caption{Comparison of different training data sources on AMBER benchmark. All methods use 10k training samples.}
\label{tab:onpolicy1}
\end{table}

\paragraph{Analysis of online Learning}

To validate the effectiveness of on-policy learning, we compare three training strategies using 10k samples: (1) SFT with Qwen3-VL-8B-Instruct distilled data, (2) SFT with LLaVA's own generated data, and (3) SFT with chosen samples from our OSCAR-constructed preference data. Results are shown in Tab.~\ref{tab:onpolicy1}. SFT with Qwen3-VL distilled data increases CHAIR from 7.6 to 9.2 and Hal from 31.2 to 62.7, confirming the Supervision-Perception Mismatch discussed in Section~\ref{3.2}. SFT with LLaVA's own generated data maintains similar performance but yields no improvement. In contrast, SFT with our OSCAR-constructed data substantially reduces CHAIR to 4.5, Hal to 15.4, and Cog to 1.4, demonstrating that our MCTS-guided data construction effectively leverages the model's discriminative capability to generate high-quality online training data.



\paragraph{Case Study}

\begin{figure}[t]
\centering
\includegraphics[width=\linewidth]{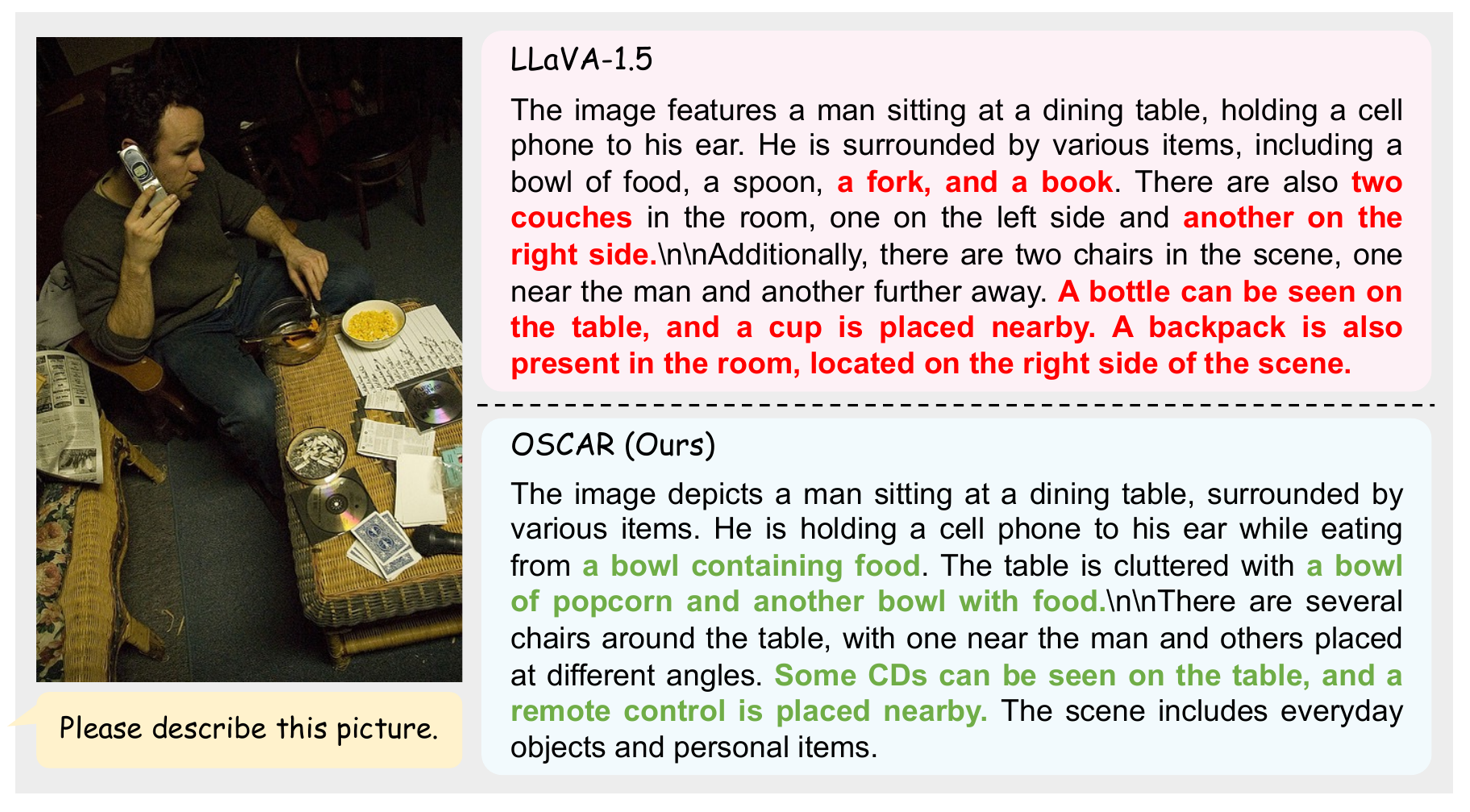}
\caption{Qualitative comparison between LLaVA-1.5 and OSCAR. Hallucinated content is highlighted in \textcolor{red}{red}, while correct descriptions are shown in \textcolor{green}{green}. OSCAR generates fewer hallucinations.}
\label{fig:case_study}
\end{figure}

Fig.~\ref{fig:case_study} presents a qualitative comparison between LLaVA-1.5 and our OSCAR method. LLaVA-1.5 generates numerous hallucinated objects (highlighted in red) that are entirely absent from the image, while in contrast, our OSCAR method produces significantly fewer hallucinations. Moreover, the response generated by OSCAR exhibits better fluency and reduced redundancy, resulting in a more concise and coherent description. This demonstrates that our dual-granularity reward mechanism effectively balances visual faithfulness with overall response quality.

\section{Conclusion}

In this paper, we identified two key observations: a Supervision-Perception Mismatch in offline preference learning that unexpectedly worsen hallucinations, and a Generative-Discriminative Gap that provides reliable self-supervision signals. Building on these insights, we proposed Iterative Self-Calibration (OSCAR), which integrates MCTS with a Dual-Granularity Reward Mechanism for online preference learning. 
Experiments demonstrated that OSCAR achieves state-of-the-art performance on hallucination benchmarks while improving general multimodal capabilities. 
Our work highlights the importance of respecting the model's intrinsic perceptual boundaries
, offering a new perspective for building more reliable vision-language systems.

\bibliographystyle{named}
\bibliography{ijcai26}

\appendix




\end{document}